# Handwriting and Drawing for Depression Detection: A Preliminary Study

Gennaro Raimo[1], Michele Buonanno[1], Massimiliano Conson[1], Gennaro Cordasco[1], Marcos Faundez-Zanuy[2], Stefano Marrone[3], Fiammetta Marulli[3], Alessandro Vinciarelli[4], and Anna Esposito[1]

[1] University of Campania "Luigi Vanvitelli", Department of Psychology, Viale Ellittico 31, 81100 Caserta, Italy. {gennaro.raimo, massimiliano.conoson, gennaro.cordasco, anna.esposito}@unicampania.it
[2] TecnoCampus Mataro-Maresme Escola Superior Politecnica Mataro, Carrer d'Ernest Lluch 32, 08302 Mataró, Barcelona, Spain. faundez@tecnocampus.cat
[3] University of Campania "Luigi Vanvitelli", Department of Mathematics, Viale Lincoln 5, 81100 Caserta, Italy. {stefano.marrone, fiammetta.marulli}@unicampania.it
[4] University of Glasgow, School of computing science, 18 Lilybank Gardens, G12 8RZ Glasgow, Scotland. alessandro.vinciarelli@glasgow.ac.uk

**Abstract.** The events of the past 2 years related to the pandemic have shown that it is increasingly important to find new tools to help mental health experts in diagnosing mood disorders. Leaving aside the long-covid cognitive (e.g., difficulty in concentration) and bodily (e.g., loss of smell) effects, the short-term covid effects on mental health were a significant increase in anxiety and depressive symptoms. The aim of this study is to use a new tool, the "online" handwriting and drawing analysis, to discriminate between healthy individuals and depressed patients. To this aim, patients with clinical depression (n = 14), individuals with high sub-clinical (diagnosed by a test rather than a doctor) depressive traits (n = 15) and healthy individuals (n = 20) were recruited and asked to perform four online drawing /handwriting tasks using a digitizing tablet and a special writing device. From the raw collected online data, seventeen drawing/writing features (categorized into five categories) were extracted, and compared among the three groups of the involved participants, through ANOVA repeated measures analyses. Results shows that Time features are more effective in discriminating between healthy and participants with sub-clinical depressive characteristics. On the other hand, Ductus and Pressure features are more effective in discriminating between clinical depressed and healthy participants.

## 1 Introduction

In recent years, there has been a significant increase of mood disorders (like depression and anxiety): an increase respectively of 18.4% and 14.9% in depres-sion and anxiety cases took place between 2005 and 2015. According to data reported by Mental Health America [17](https://mhanational.org/), from January to September 2020, there was an increase, compared to 2019, of 62% in the request for screening for depression and of 93% in the request for screening to assess anxiety. In addition, the severity of anxiety and depressive symptoms also increased significantly, with about 80% of screened people showing severe anxiety and depression symptoms [17]. This data demonstrates that the extraordinary, stressful pandemic situation had a very strong impact on the emotional life of the worldwide population. Furthermore, this increase in cases shows how much more important is to find new tools to detect the presence (or the absence) of these types of disorders quickly and effectively.

Mood disorders (also called mood affective disorders) are characterized by cognitive (memory and concentration difficulties, worry/rumination, guilt, self-depreciation), emotional (low mood, sadness, distress, irritability, hopeless), physical (loss of energy, fatigue, weight loss or gain, anhedonia, sleep disturbances) and behavioral (psychomotor retardation, social withdrawal, avoidance, loss of interest in enjoyable activities, suicide thoughts and suicide acts) symptoms [6].

Therefore, the detection of these diseases can be based on the analysis of behavioral signals (speeches, body motions, physiological data, handwritings, text analysis and drawings; for a review see [8]), which have been proven to be efficient, effective and have no subjective bias.

Specifically, one of these, handwriting, proved useful to identify people's different characteristics: socio-personal data like age, nationality and gender [1, 3]: but also, personality traits [7, 9]; neurodegenerative pathology [23] and, even more, emotional states [4, 12]. Considering these results and considering that handwriting and drawing skills are common to most humans, it is conceivable to think that is possible to identify and discriminate depression patients using these two skills.

Several studies have demonstrated, using a kinematic approach (that analyze movements), that healthy participants are faster [13, 14] and more accurate [21] than depressed patients. However, these studies are limited because they analyzes only a specific group or a specific task. Regarding participants, they involved only elderly people [21] or healthy young (students between 18-22 years old) people [13]; instead, for what concerns the tasks, the limits were the use of uncommon and non-naturalistic tasks (i.e. drawing of a circle) [14, 21] or a stress load test [13]. Rosenblum [20], still using a kinematic approach, found that pressure analysis could help discriminate between depressed patients and non-clinical participants. In addition, in this case, the study was focused only on elderly people.

Unlike the previous studies, Gawda [10] found no statistically significant differences in velocity comparing depressed patients, borderline patients, and healthy participants. She found only more tremors and a more descending trait. Results of these studies are heterogeneous, showing that there is still clarity to be done in this field.

The use of new tools to discover, discriminate and identify these diseases is also necessary because recent studies [16, 24] and even the European community, in last years highlighted the importance of prevention programs for early detection of depression and anxiety. Early diagnosis leads to greater efficacy of therapy and faster improvement in patients' overall health.

At author knowledge, literature shows that no studies have been conducted with comparison between healthy and clinical participants using our approach and our task. The present paper is a pilot study that reports a comparison of these types of participants (clinical and healthy) considering "online" handwriting and drawing skills to evaluate clinical and typical participants' differences.

## 2   Handwriting analysis for the detection of depression and anxiety

New technologies to support handwriting analysis have made it possible thanks to tools that can analyze "online" writing and drawing skills. This has been possible thanks to the development of pencil and digital tablets, which can analyze parameters such as pressure applied on paper, inclination of the pen used, number of strokes used to complete a task, etc.

An INTUOS WACOM series 4 digital tablet was used to collect data. Since the standard pen did not allow writing on a common paper, so participants can't see their stroke in the way they are used to, another pen (which allows to see strokes) named Intuos Inkpen, was paired with the tablet. In this way, it was possible to obtain information both on the strokes (through the sheet of paper), and on other parameters, as time information and in-air movements (through the tablet).

Data was collected for several task and recorded in a comma separated values (CSV) file. The system registered the following data every 8ms:

- the two-dimensional coordinates of the pen;
- the unix epoch (the number of milliseconds that have elapsed since January 1, 1970;
- the pen status (on paper=1 or in-air=0);
- two angles that describes the position of the paper with respect to the tablet;
- the amount of pressure applied by the pen on the paper, expressed using an integer value, from 0 (no pressure) to 255 (maximum pressure). the amount of pressure applied by the pen on the paper, expressed using an integer value, from 0 (no pressure) to 255 (maximum pressure).

## 3   Participants and methodology

### 3.1   Participants

The experiments involved a total of 49 participants (Male = 15; mean age = 28.90; SD = 10.47) partitioned as follows:

- 14 clinical group (depression patients)
- 15 DASS-Severe Group
- 20 DASS-Normal Group

Description of socio-personal characteristics (group, mean age and standard deviation of age) are reported in the table below (Tab. 1).

**Table 1.** Group participants with sample size (N), mean of age (M) and standard deviation of age (SD). CL = Clinical Group, SEV = DASS-Severe Group; NOR = DASS-Normal Group.

| Group | N | M | SD |
|-------|----|-------|-------|
| CL | 14 | 39,86 | 14,40 |
| SEV | 15 | 24,53 | 2,72 |
| N | 20 | 24,50 | 2,42 |

Healthy participants, recruited at Università degli Studi della Campania "L. Vanvitellli", were included only if satisfying the following inclusion criteria: 1. right-handed; 2. no clinical diagnosis of neurological, neuropsychiatric, or psychological disease; 3a. for DASS-Normal group, "normal" score in Depression Anxiety Stress (DASS) questionnaire, in all the three sub-scale (depression, anxiety, stress); 3b. for DASS-Severe group, "severe" or "extremely severe" score in DASS questionnaire in depression sub-scale.

The DASS questionnaire [2] is a self-report measure that estimate affective states of depression, anxiety, and stress. Each scale is composed of seven items (21 total items). Responses are given on a 4-point Likert scale (from 0: Did not apply to me at all; to 4: Applied to me very much or most of the time). The scores obtained for each subscale indicate the severity of the symptoms from normal (healthy individual), to mild, moderate, severe, and extremely severe condition.

The participants of the clinical group are patients with clinical depression, diagnosed by a mental health-care expert (a psychologist or a psychiatrist) with a formal diagnosis. The clinical group participants were also right-handed. All the participants speak Italian as their native language.

### 3.2  Tasks and Features

The participants in the clinical group and healthy groups completed 2 different protocols (see Fig. 1), which share four drawing or handwriting tasks:

1. drawing of two pentagons;
2. drawing of a house;
3. writing an Italian sentence in cursive letters;
4. drawing of a clock with hours and clock hands.

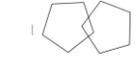

**Fig. 1.** (left) clinical group protocol; (right) healthy group protocol (filled by one participant). Four tasks are common.

The "online" data acquisition allows us to obtain detailed information on parameters of the written strokes, such as trajectory, speed, and time features. In addition, this tool also enables us to detect features that have proved useful in this field, such as in-air movements (movements of the pen close to the paper) [19, 22].

For this study we separated the traits into three categories, as described in [3] and [4]:

- in-air (but close to the paper), pen status 0;
- on-paper, pen status 1;
- idle (far from the paper), not recorded but recognizable using timestamps.

In particular, we considered 17 features (for a description of these see [3]), divided into five categories:

a) Pressure: a value of the pressure applied by the pen on the paper during a specific task (considering only on-paper traits);
   1. Pmin, the lowest pressure value applied;
   2. Pmax, the highest pressure value applied;
   3. Pavg, the average value of the applied pressure;

    4. Psd, the standard deviation of the applied pressures;
    5. P10, lower 10th percentile of applied pressures;
    6. P90, lower 90th percentile of applied pressures.
b) Ductus: the number of strokes in each pen status;
    7. Nup, amount of in-air traits;
    8. Ndown, amount of on-paper traits;
    9. Nidle, total amount of traits not recognized by the tablet.
c) Time: total time spent, in each pen status, to complete the task;
    10. Tup, time of in-air pen status;
    11. Tdown, time of on-paper pen status;
    12. Tidle, time of pen status not recognized by the tablet;
    13. Ttotal = Tup + Tdown + Tidle, total time used to complete the task.
d) Space: the occupied space by the strokes task (considering exclusively on-paper traits);
    14. Sbb, a value obtained computing the smallest axis aligned bounding box containing the stroke and summing its area;
    15. Savg, average lengths of empty spaces between consecutive strokes;
    16. Stotal, sum of the lengths of empty spaces between consecutive on-paper strokes.
e) Inclination; a value of strokes inclination.
    17. Iavg, average inclination of the diagonals of the bounding boxes containing the strokes.

## 4  Results

Several repeated measures ANOVA analyses were performed to evaluate the efficacy of the considered features in differentiating the groups of writers. We carried out an ANOVA for each task and for each features category. Furthermore, the group (clinical / healthy) of the participants was considered as between subjects' factors while scoring of participants on each feature was considered as within subjects factors. We set at $α < .05$ the significance level; finally, Bonferroni's post hoc tests was conducted to assess differences among means.

Below we provide the significant results for each feature category. Raw data and all statistical results are available upon request (gennaro.raimo@unicampania.it).

### 4.1  Ductus

For the ductus category, the results show that emerges a significant difference between the groups, with differences between the clinical patient group and healthy groups. From the descriptive data it is possible to notice how the number of traits increases according to the worsening of the depressive condition. Table 2 shows a description of scores for all groups and all tasks. From the pairwise comparisons, in all the four tasks considered a difference emerges between the clinical group and the DASS-Normal group. Furthermore, in tasks 2-3-4 there is a difference between the clinical group and DASS-Severe group. Finally, in task 1, a difference emerges between the DASS-Severe and DASS-Normal groups.

Statistical data divided by task is shown below:

a) Task 1: Drawing of two-pentagon: a significant difference emerges between the three groups [$F$ (2; 46) = 13.222; $p \ll .01$]. Clinical group significantly differs from the) Task 1: Drawing of two-pentagon: a significant difference DASS-Normal group ($p \ll .01$). Moreover, DASS-Normal group significantly differs from the DASS-Severe group ($p = .031$).
b) Task 2: Drawing of a house: a significant difference emerges between the three groups [$F$ (2; 46) = 19.347; $p \ll .01$]. Clinical group significantly differs from the DASS-Normal group ($p \ll .01$). Moreover, Clinical group significantly differs from the DASS-Severe group ($p \ll .01$).
c) Task 3: Writing of Italian sentence in cursive letters: a significant difference emerges between the three groups [$F$ (2; 46) = 18.090; $p \ll .01$]. Clinical group significantly differs from the DASS-Normal group ($p \ll .01$). Moreover, Clinical group significantly differs from the DASS-Severe group ($p < .01$).
d) Task 4: Drawing a clock with hours and hands: a significant difference emerges between the three groups [$F$ (2; 46) = 18.464; $p \ll .01$]. Clinical group significantly differs from the DASS-Normal group ($p \ll .01$). Moreover, Clinical group significantly differs from the DASS-Severe group ($p \ll .01$).

Analyzing individually the different parameters of ductus category, the following differences emerge between the groups:

- Nup: Clinical group and DASS-Normal group (all tasks: $p \ll .01$).
- Nup: Clinical group and DASS-Severe group (task 2-3-4: $p \ll .01$).
- Ndown: Clinical group and DASS-Normal group (all tasks: $p \ll .01$).
- Ndown: Clinical group and DASS-Severe group (task 1: $p = .045$; task 2-3-4: $p \ll .01$).
- Nidle: Clinical group and DASS-Severe group (task 2: $p \ll .01$).

### 4.2  Time

For the time category, the results show that emerges a significant difference between the groups. Also in this case, observing descriptive data it is possible to notice how time taken to perform a task is conditioned by the depressive situation in progress, with times that increase when the depressive state is worst. Description of scoring separately for tasks and groups are reported in table 3 (only task with significantly differences is reported). From the pairwise comparisons, in task 1 a significant difference emerges between the DASS-Normal group and the DASS-Severe group; instead, in task 3 a significant difference emerges between the clinical group and DASS-Normal group. No other significant differences emerged in the tasks considered in this research.

**Table 2.** Performance (number of strokes) on the ductus category, separately for group and task. The values are expressed as mean (standard deviations). CL = Clinical Group, SEV = DASS-Severe Group, NOR = DASS-Normal Group

| TASK | DUCTUS CATEGORY | | |
|---|---|---|---|
| | Group | | |
| | CL | SEV | NOR |
| Task 1 | 26.286 (3.480) | 14.978 (3.362) | 3.100 (2.912) |
| Task 2 | 49.857 (4.167) | 24.600 (4.304) | 16.600 (3.494) |
| Task 3 | 47.643 (3.665) | 25.111 (3.541) | 19.733 (3.066) |
| Task 4 | 60.214 (3.927) | 35.578 (3.794) | 30.083 (3.286) |

Statistical data divided by task is shown below:

a) Task 1: Drawing of two-pentagon: a significant difference emerges between the three groups [$F$ (2; 46) = 4.068; $p$ = .024]. DASS-Normal significantly differs from the DASS-Severe group ($p$ = .045).
b) Task 3: Writing of Italian sentence in cursive letters: a significant difference emerges between the three groups [$F$ (2; 46) = 6, 580; $p$ = .003]. Clinical group significantly differs from the DASS-Normal group ($p$ = .002).

**Table 3.** Performance (time spent in millisecond) on the time category, separately for group and task. The values are expressed as mean (standard deviations). CL = Clinical Group, SEV = DASS-Severe Group, NOR = DASS-Normal Group.

| TASK | TIME CATEGORY | | |
|---|---|---|---|
| | Group | | |
| | CL | SEV | NOR |
| Task 1 | 17675.821 (1556.496 | 12131.050 (1503.718) | 7096.075 (1302.258) |
| Task 2 | 23026.286 (2020.529) | 18031.850 (1952.016) | 13494.300 (1690.496) |

Analyzing individually the different parameters of time category, the following differences emerge between the groups:

- Tup: Clinical group and DASS-Normal group (task 1: $p$ = .002; task 3: $p$ = .001)
- Tup: DASS-Normal group e DASS-Severe group (task 1: $p$ = .039)
- Tdown: Clinical group and DASS-Normal group (task 3: $p$ = .01)
- Tidle: DASS-Normal group e DASS-Severe group (task 1: $p$ = .032)
- Ttotal: Clinical group and DASS-Normal group (task 3: $p$ = .002)
- Ttotal: DASS-Normal group e DASS-Severe group (task 1: $p$ = .045)

### 4.3 Pressure

Also for the pressure category, the results show that emerges a significant difference between the groups. Unlike the other two categories, the pressure applied on the paper tends to decrease with the worsening of the depressive state. Looking at the descriptive data it is possible to highlight how in healthy individuals the pressure applied is greater. From the pairwise comparisons, in tasks 1 and 2 a difference emerges between the clinical group and the DASS-Normal group. No significant differences emerged in the other tasks considered in this research. Table 4 described scoring obtained in pressure category separately for groups and tasks (only task with significantly differences is reported).

Statistical data divided by task is shown below:

a) Task 1: Drawing of two-pentagon: a significant difference emerges between the three groups [$F(2; 46) = 9.101; p \ll .01$]. Clinical group significantly differs from the DASS-Normal group ($p \ll .01$).
b) Task 2: Drawing of a house: a significant difference emerges between the three groups [$F(2; 46) = 4.555; p = 0.016$]. Clinical group significantly differs from the DASS-Normal group ($p = .020$).

**Table 4.** Performance (pressure applied by the pen on the paper) on the pressure category, separately for group and task. The values are expressed as mean (standard deviations). CL = Clinical Group; SEV = DASS-Severe Group, NOR = DASS-Normal Group.

| TASK | PRESSURE CATEGORY | | |
|---|---|---|---|
| | Group | | |
| | CL | SEV | NOR |
| Task 1 | 369.646 (22.644) | 446.241 (21.876) | 495.572 (18.945) |
| Task 2 | 339.659 (23.303) | 417.886 (22.513) | 426.029 (19.497) |

Analyzing individually the different parameters of Pressure category, the following differences emerge between the groups:

- Pmax: Clinical group and DASS-Normal group (task 1: $p = .015$).
- Pmin: Clinical group and DASS-Normal group (task 2: $p = .003$).
- Pavg: Clinical group and DASS-Normal group (task 1: $p \ll ..01$; task 2: $p = .018$)
- P10: Clinical group and DASS-Normal group (task 1: $p \ll .01$; task 2: $p = .001$)
- P10: Clinical group and DASS-Severe group (task 1: $p = .021$; task 2: $p = .019$)
- P90: Clinical group and DASS-Normal group (task 1: $p = .002$; task 2: $p = .042$)

## 5     Discussion

The most recent global burden of disease study [11] estimated that depression, from 1990 to 2017, ranks third among those responsible causes worldwide of years lived with disabilities (YLD). Furthermore, depression also leads the ranking of suicides cause, with estimated 800.000 victims worldwide; moreover, it is the second leading cause of death between 15-29 years [5].

Experiencing adverse life events, like a pandemic situation or a social isolation lockdown, can increase the likelihood of developing depression, also after many years [15]. Considering that catastrophic events (i.e. pandemic situation or a social lockdown) could happen again in the next years, it is essential to develop new tools to recognize and identify depression early and quickly.

Other studies have already shown how it is possible to discriminate between pathological/sub-clinical patients and healthy participants through behavioral analyses [4,8,10]; not only with mood disorders but also with Alzheimer's disease (AD) and of related dementias [18].

In this way, this study demonstrates that it is possible to discriminate between depressed clinical patients and non-clinical participants, exploiting online handwriting and drawing features, obtained from specific tasks such as the drawing of: two-pentagons (task 1), a house (task 2), and a clock with the twelve hours and clock hands (task 4), as well as the writing in cursive letters (task 3), focusing on features derived from Ductus, Pressure and Timing measures.

Specifically, timing measures are effective in discriminating between healthy individuals and participants with sub-clinical depressive traits. Instead, ductus and pressure measures are well suited for discriminating between healthy and clinical depressed participants.

The limitations of this study are evidently a relatively small number of participants and the greater number of women than men in the sample.

Future studies aim to find more evidence on larger samples of subjects, even better if matched by gender and age. Moreover, machine learning techniques can be used to analyze the data to increase the reliability and validity of this experimental paradigm and to have a classification accuracy. Lastly, future goals can also be considered to arrive at a complete automatic detection of mood disorders and improve in a clinical field all these advances.

## 6     Conclusion

Considering results of this study, handwriting analysis appears to be a promising tool for speeding the diagnosis of depression, providing quantitative measures and decreasing the health care costs associated to long professional interviews.

Furthermore, this approach can offer more reliable and effective results, since it is not linked to subjective evaluation bias, which can be present both in patients and in mental-health experts and could affect the diagnosis.

In conclusion, the aims of this study on handwriting and drawing skills for the detection of depression are: clarify the discordant results present in the literature regarding this topic, to have a specific, effective and quickly and especially standardized methodology for this type of evaluation; rise the interests of scientific community and to all researchers interested in this field of study.

**Acknowledgment** The research leading to these results has received funding from the project ANDROIDS funded by the program V:ALERE 2019 Università della Campania "Luigi Vanvitelli", D.R. 906 del 4/10/2019, prot. n. 157264, 17/10/2019, the project SIROBOTICS that received funding from Italian MIUR, PNR 2015-2020, D.D. 1735, 13/07/ 2017, and the EU H2020 research and innovation program under grant agreement N. 769872 (EMPATHIC) and N. 823907 (MENHIR).